\documentclass[sigconf]{aamas}  % do not change this line!

\AtBeginDocument{%
  \providecommand\BibTeX{{%
    \normalfont B\kern-0.5em{\scshape i\kern-0.25em b}\kern-0.8em\TeX}}}
    
\usepackage{booktabs}    
\usepackage{flushend} % do not change this line!
\usepackage[ruled]{algorithm2e}

\setcopyright{ifaamas}  % do not change this line!
\copyrightyear{2020} % do not change this line!
\acmYear{2020} % do not change this line!
\acmDOI{} % do not change this line!
\acmPrice{} % do not change this line!
\acmISBN{} % do not change this line!
\acmConference[AAMAS'20]{Proc.\@ of the 19th International Conference on Autonomous Agents and Multiagent Systems (AAMAS 2020)}{May 9--13, 2020}{Auckland, New Zealand}{B.~An, N.~Yorke-Smith, A.~El~Fallah~Seghrouchni, G.~Sukthankar (eds.)}  % do not change this line!

\newcommand{\tb}[1]{\textup{\textbf{#1}}}
\newcommand{\tu}[1]{\textup{#1}}
\newcommand{\p}[1]{{#1}^\prime}

\begin{document}

\title{Deep Residual Reinforcement Learning}
% \title{Sample AAMAS Paper using the New ACM LaTeX Template}  % put your title here!
% \titlenote{Titlenote --- e.g., used for things like "This article extends an earlier paper titled XYZ", and "equal contribution by the first two authors".}

% AAMAS: as appropriate, uncomment one subtitle line; see camera ready instructions
%\subtitle{Extended Abstract}
%\subtitle{Blue Sky Ideas Track}
%\subtitle{JAAMAS Track}
%\subtitle{Doctoral Consortium}                              
%\subtitle{Demonstration}
%\subtitlenote{Please refrain from using subtitle notes}

\author{Shangtong Zhang, Wendelin Boehmer, Shimon Whiteson}
\affiliation{Department of Computer Science, University of Oxford, United Kingdom}
% \email{shangtong.zhang@cs.ox.ac.uk, wendelin.boehmer@cs.ox.ac.uk, shimon.whiteson@cs.ox.ac.uk}
\email{shangtong.zhang@cs.ox.ac.uk}

\begin{abstract}  % put your abstract here!
    We revisit \emph{residual algorithms} in both model-free and model-based reinforcement learning settings. 
    We propose the \emph{bidirectional target network} technique to stabilize residual algorithms, yielding a residual version of DDPG that significantly outperforms vanilla DDPG in the DeepMind Control Suite benchmark. 
    Moreover, we find the residual algorithm an effective approach to the \emph{distribution mismatch} problem in model-based planning. 
    Compared with the existing TD($k$) method, our residual-based method makes weaker assumptions about the model and yields a greater performance boost.
\end{abstract}

% AAMAS: the ACM CCS are encouraged but optional within AAMAS papers
%\begin{CCSXML}
%<ccs2012>
%  <concept_id>10010520.10010553.10010562</concept_id>
%  <concept_desc>Computer systems organization~Embedded systems</concept_desc>
%  <concept_significance>500</concept_significance>
%  <concept_id>10010520.10010575.10010755</concept_id>
%  <concept_desc>Computer systems organization~Redundancy</concept_desc>
%  <concept_significance>300</concept_significance>
%  <concept_id>10010520.10010553.10010554</concept_id>
%  <concept_desc>Computer systems organization~Robotics</concept_desc>
%  <concept_significance>100</concept_significance>
%  <concept_id>10003033.10003083.10003095</concept_id>
%  <concept_desc>Networks~Network reliability</concept_desc>
%  <concept_significance>100</concept_significance>
%</ccs2012>
%\end{CCSXML}
%
%\ccsdesc[500]{Computer systems organization~Embedded systems}


\maketitle

%%%%%%%%%%%%%%%%%%%%%%%%%%%%%%%%%%%%%%%%%%%%%%%%%%%%%%%%%%%%%%%%%%%%%%%%%%%%%%%%%%%%%%%%%%%%%%%%%%%%%%%%%
%% start of main body of paper

% \input{samplebody-conf}
\section{Introduction}
Semi-gradient algorithms have recently enjoyed great success in deep reinforcement learning (RL) problems, e.g., DQN \citep{mnih2015human} achieves human-level control in the Arcade Learning Environment (ALE, \cite{bellemare2013arcade}). 
However, such algorithms lack theoretical support.
Most semi-gradient algorithms suffer from divergence under nonlinear function approximation or off-policy training \citep{baird1995residual,tsitsiklis1997analysis}.
By contrast, \emph{residual gradient} (RG, \cite{baird1995residual}) algorithms are true stochastic gradient algorithms and enjoy convergence guarantees (to a local minimum) under mild conditions with both nonlinear function approximation and off-policy training. 
\citet{baird1995residual} further proposes \emph{residual algorithms} (RA) to unify residual gradients and semi-gradients via mixing them together.

Residual algorithms suffer from the double sampling issue \citep{baird1995residual}: two independently sampled successor states are required to compute the residual gradients. 
This requirement can be easily satisfied in model-based RL or in deterministic environments.
However, even in these settings, residual algorithms have long been either overlooked or dismissed as impractical.
In this paper, we aim to overturn that conventional wisdom with new algorithms built on RA and empirical results showing their efficacy. 

Our contributions are threefold. 
First, we give a thorough overview of existing comparisons between residual gradient algorithms and semi-gradient algorithms. 

Second, we showcase the advantages of RA in a model-free RL setting with deterministic environments. 
While target networks \citep{mnih2015human} are usually an important component in deep RL algorithms to stabilize training \citep{mnih2015human,lillicrap2015continuous},
we find a naive combination of target networks and residual algorithms, in general, does not improve performance.  Therefore, we propose the \emph{bidirectional target network} technique to stabilize residual algorithms. 
We show that our residual version of Deep Deterministic Policy Gradients (DDPG, \cite{lillicrap2015continuous}) significantly outperforms vanilla DDPG in the DeepMind Control Suite (DMControl, \cite{tassa2018deepmind}) and Mujoco benchmarks. 

Third, we showcase the advantages of RA in a model-based RL setting, where a learned model generates imaginary transitions to train the value function. 
In general, model-based methods suffer from a \emph{distribution mismatch} problem \citep{feinberg2018model}.
The value function trained on real states does not generalize well to imaginary states generated by a model. 
To address this issue, \citet{feinberg2018model} train the value function on both real and imaginary states via the TD($k$) trick. 
However, TD($k$) requires that predictions $k$ steps in the future made by model rollouts will be accurate \citep{feinberg2018model}.
In this paper, we show that RA naturally allows the value function to be trained on both real and imaginary states and requires only 1-step rollouts.
Our experiments show that RA-based planning boosts performance more than TD($k$)-based planning in most cases. 

\section{Background}

We consider an MDP \citep{puterman2014markov} consisting of a finite state space $\mathcal{S}$, a finite action space $\mathcal{A}$, a reward function $r: \mathcal{S} \times \mathcal{A} \rightarrow \mathbb{R}$, a transition kernel $p: \mathcal{S} \times \mathcal{S} \times \mathcal{A} \rightarrow [0, 1]$ and a discount factor $\gamma \in [0, 1)$. 
With $\pi: \mathcal{A} \times \mathcal{S} \rightarrow [0, 1]$ denoting a policy,
at time $t$, an agent at a state $S_t$ takes an action $A_t$ according to $\pi(\cdot | S_t)$. 
The agent then gets a reward $R_{t+1}$ satisfying $\mathbb{E}[R_{t+1}] = r(S_t, A_t)$ and proceeds to a new state $S_{t+1}$ according to $p(\cdot | S_t, A_t)$.
We use $G_t \doteq \sum_{i=t+1}^\infty \gamma^{i-t-1}R_i$ to denote the return from time $t$,
$v_\pi(s) \doteq \mathbb{E}_\pi[G_t \mid S_t = s]$ to denote the state value function of $\pi$,
and $q_\pi(s, a) \doteq \mathbb{E}_\pi[G_t \mid S_t = s, A_t = a]$ to denote the state-action value function of $\pi$. 
In the rest of this section, we use a bold capital letter to denote a matrix and a bold lowercase letter to denote a column vector. 
We use $\tb{P}_\pi$ to denote the transition matrix induced by a policy $\pi$, i.e., $\tb{P}_\pi[s, \p{s}] \doteq \sum_a \pi(s, a) p(\p{s}|s, a)$, 
and use $d_\pi$ to denote its unique stationary distribution,
assuming $\tb{P}_\pi$ is ergodic.
The reward vector induced by $\pi$ is $\tb{r}_\pi[s] = \sum_a \pi(a | s) r(s, a)$.

The value function $v_\pi$ is the unique fixed point of the Bellman operator $\mathcal{T}$ \citep{bellman2013dynamic}.
In a matrix form, $\mathcal{T}$ is defined as $\mathcal{T}\tb{v} \doteq \tb{r}_\pi+\gamma \tb{P}_\pi \tb{v}$, where $\tb{v}$ can be any vector in $\mathbb{R}^N$. 
Here $N \doteq |\mathcal{S}|$ is the number of states.

\textbf{Policy Evaluation:} 
We consider the problem of finding $v_\pi$ for a given policy $\pi$ and use $\tb{v}$, parameterized by $\tb{w} \in \mathbb{R}^d$, to denote an estimate of $\tb{v}_\pi$, the vector form of $v_\pi$. 
We start with on-policy linear function approximation
and use $x: \mathcal{S} \rightarrow \mathbb{R}^d$ to denote a feature function which maps a state to a $d$-dimensional feature.
The feature matrix is then $\tb{X} \doteq [\tb{x}(s_1), \dots, \tb{x}(s_N)]^\textup{T} \in \mathbb{R}^{N \times d}$, 
and the value estimate is $\tb{v} \doteq \tb{X}\tb{w}$. 

To approximate $\tb{v}_\pi$, one direct goal is to minimize the Mean Squared Value Error:
\begin{align*}
\textstyle
\tu{MSVE}(\tb{w}) \doteq ||\tb{v} - \tb{v}_\pi||^2_{d_\pi} \doteq \sum_s d_\pi(s) \big(\tb{v}(s) - \tb{v}_\pi(s)\big)^2.
\end{align*}
To minimize MSVE, a Monte Carlo return can be used as a sample for $\tb{v}_\pi$ to train $\tb{v}$. 
However, this method suffers from a large variance and usually requires off-line learning \citep{bertsekas1996neuro}. 
To address those issues, we consider minimizing the Mean Squared Projected Bellman Error (MSPBE) and the Mean Squared Bellman Error (MSBE):
\begin{align*}
\tu{MSPBE}(\tb{w}) \doteq ||\tb{v} - \Pi\mathcal{T}\tb{v}||^2_{d_\pi}, \quad \tu{MSBE}(\tb{w}) \doteq ||\tb{v} - \mathcal{T}\tb{v}||^2_{d_\pi}.
\end{align*}
Here $\Pi$ is a projection operator which maps an arbitrary vector onto the column vector space of $\tb{X}$, minimizing a $d_\pi$-weighted projection error, i.e.,
$\Pi\tb{v} \doteq \tb{X}\bar{\tb{w}}$, where $\bar{\tb{w}} \doteq \arg\min_\tb{w} ||\tb{v} - \tb{X}\tb{w}||^2_{d_\pi}$.
With linear function approximation and fixed features, $\Pi$ is linear.

There are various algorithms for minimizing MSPBE and MSBE. 
Temporal Difference learning (TD, \cite{sutton1988learning}) is commonly used to minimize MSPBE. TD updates $\tb{w}$ as 
\begin{align*}
\tb{w} \leftarrow \tb{w} + \alpha \big(R_{t+1} + \gamma \tb{v}(S_{t+1}) - \tb{v}(S_t) \big) \nabla_\tb{w}\tb{v}(S_t),
\end{align*}
where $\alpha$ is a step size. 
Under mild conditions, on-policy linear TD converges to the point where MSPBE is 0 \citep{tsitsiklis1997analysis}. 
TD is a \emph{semi-gradient} \citep{sutton2018reinforcement} algorithm in that it ignores the dependency of $\tb{v}(S_{t+1})$ on $\tb{w}$. 
There are also \emph{true gradient} algorithms for optimizing MSPBE, e.g., Gradient TD methods \citep{sutton2009fast}. 
Gradient TD methods compute the gradient of MSPBE directly and also enjoy convergence guarantees. 

\citet{baird1995residual} proposes \emph{residual gradient} algorithms for minimizing MSBE, which updates $\tb{w}$ as 
\begin{align}
\nonumber
\label{eq:rg-td}
\tb{w} \leftarrow \tb{w} - \alpha \big(R_{t+1} + \gamma \tb{v}(S_{t+1}) - \tb{v}(S_t) \big) \\
\cdot \big(\gamma \nabla_\tb{w}\tb{v}(S^\prime_{t+1}) - \nabla_\tb{w}\tb{v}(S_t) \big),
\end{align}
where $S_{t+1}^\prime$ is another sampled successor state for $S_t$, independent of $S_{t+1}$. 
This requirement for two independent samples is known as the \emph{double sampling issue} \citep{baird1995residual}. 
If both the transition kernel $p$ and the policy $\pi$ are deterministic, we can simply use one sample without introducing bias. 
Otherwise, we may need to have access to the transition kernel $p$, which is usually not available in model-free RL. 
Regardless, RG is a true gradient algorithm with convergence guarantees under mild conditions.

We now expand our discussion about policy evaluation into  off-policy learning and nonlinear function approximation,
where the states $\{S_t\}$ are drawn according to a behavior policy $\mu$ instead of the target policy $\pi$.
True gradient algorithms like Gradient TD methods and RG remain convergent to local minima under off-policy training with any function approximator \citep{baird1995residual,sutton2009fast,maei2011gradient}. 
However, the empirical success of Gradient TD methods is limited to simple domains due to its large variance \citep{sutton2016emphatic}.
Semi-gradient algorithms are not convergent in general, e.g., the divergence of off-policy linear TD is well-documented \citep{tsitsiklis1997analysis}. 

Semi-gradient algorithms are fast but in general not convergent. 
Residual gradient algorithms are convergent but slow \citep{baird1995residual}.
To take advantage of both, 
\citet{baird1995residual} proposes to mix semi-gradients and residual gradients together, yielding the \emph{residual algorithms}. The RA version of TD \citep{baird1995residual} updates $\tb{w}$ as
\begin{align*}
\tb{w} \leftarrow \tb{w} - \alpha \big(R_{t+1} + \gamma \tb{v}(S_{t+1}) - \tb{v}(S_t) \big) \\
\cdot \big(\gamma \eta \nabla_\tb{w}\tb{v}(S^\prime_{t+1}) - \nabla_\tb{w}\tb{v}(S_t) \big),
\end{align*}
where $\eta \in [0, 1]$ controls how the two gradients are mixed. Little empirical study has been conducted for RA.

\textbf{Control:} We now consider the problem of control, where we are interested in finding an optimal policy $\pi^*$ such that $ v_{\pi^*}(s) \geq v_\pi(s) \, \forall (\pi, s)$. 
We use $q_*$ to denote the state-action value function of $\pi^*$ and $Q$ to denote an estimate of $q_*$, parameterized by $\theta$.
Q-learning \citep{watkins1992q} is usually used to train $Q$ and enjoys convergence guarantees in the tabular setting. 
When Q-learning is combined with neural networks, Deep-Q-Networks (DQN, \cite{mnih2015human}) update $\theta$ as 
\begin{align}
\textstyle
\label{eq:dqn}
\nonumber
\theta \leftarrow \theta + \alpha_1 (r_{t+1} + \max_a \bar{Q}(s_{t+1}, a) - Q(s_t, a_t)) \\
\cdot \nabla_\theta Q(s_t, a_t),
\end{align}
where $\alpha_1$ is a step size, $(s_t, a_t, r_{t+1}, s_{t+1})$ is a transition sampled from a replay buffer \citep{lin1992self}, 
and $\bar{Q}$ indicates the estimate is from a target network \citep{mnih2015human}, 
parameterized by $\theta^-$, which is synchronized with $\theta$ periodically.

When the action space is continuous, it is hard to perform the $\max$ operation in the DQN update~\eqref{eq:dqn}.
DDPG can be interpreted as a continuous version of DQN, 
where an actor $\mu: \mathcal{S} \rightarrow \mathcal{A}$, parameterized by $\nu$, is trained to output the greedy action.
DDPG updates $\theta$ and $\nu$ as
\begin{align}  
\label{eq:ddpg-critic}
\nonumber
\theta &\leftarrow \theta + \alpha_1 \big(r_{t+1} \\
&+ \gamma \bar{Q}(s_{t+1}, \bar{\mu}(s_{t+1})) - Q(s_t, a_t) \big) \nabla_\theta Q(s_t, a_t), \\
\label{eq:ddpg-actor}
\nu &\leftarrow \nu + \alpha_2 \nabla_a Q(s_t, a)|_{a = \mu(s_t)} \nabla_\nu \mu(s_t),
\end{align}
where $\alpha_2$ is a step size, 
$\bar{\mu}$ indicates the greedy action is from a target network, parameterized by $\nu^-$. 

Both DQN and DDPG are semi-gradient algorithms.
There are also true gradient methods for control, e.g., Greedy-GQ \citep{maei2010toward} and the residual version of Q-learning \citep{baird1995residual}.
As with Gradient TD methods, the empirical success of Greedy-GQ is limited to simple domains due to its large variance \citep{sutton2016emphatic}.

\section{Comparing TD and RG}
\label{sec:cmp}
In this section, we review existing comparisons between RG and TD. We start by comparing their fixed points, MSBE and MSPBE, in the setting of linear function approximation.

\textbf{Cons of MSBE:} 
\begin{itemize}
\item \citet{sutton2018reinforcement} show that MSBE is not uniquely determined by the observed data. 
Different MDPs may have the same data distribution due to state aliasing, 
but the minima of MSBE can still be different. 
This questions the learnability of MSBE as sampled transitions are all that is available in model-free RL. 
By contrast, the minima of MSPBE are always the same for MDPs with the same data distribution. 
\item Empirically, optimizing MSBE can lead to unsatisfying solutions. 
For example, in the A-presplit example \citep{sutton2018reinforcement}, 
the value of most states can be represented accurately by the function approximator but the MSBE minimizer does not do so, while the MSPBE minimizer does.
Furthermore, empirically the MSBE minimizer can be further from the MSVE minimizer than the MSPBE minimizer \citep{dann2014policy}.

\end{itemize}

\textbf{Pros of MSBE:} 
\begin{itemize}
\item \citet{williams1993tight} show MSBE can be used to bound MSVE (up to a constant). 
By contrast, at a point where MSPBE is minimized, MSVE can be arbitrarily large \citep{bertsekas1996neuro}.
\item MSBE is an upper bound of MSPBE \citep{scherrer2010should}, indicating that optimizing MSBE implicitly optimizes MSPBE.
\end{itemize}
We now compare RG and TD.

\textbf{Cons of RG:} 
\begin{itemize}
\item Due to the double sampling issue, it is usually hard to apply RG in the stochastic model-free setting \citep{baird1995residual}, while TD is compatible with both deterministic and stochastic environments.
\item RG is usually slower than TD. Empirically, this is observed by
\citet{baird1995residual}, \citet{van2011insights}, \citet{gordon1995stable} and \citet{gordon1999approximate}. 
Intuitively, in the RG update \eqref{eq:rg-td},  
a state $S_t$ and its successor $S_{t+1}^\prime$ are often similar under function approximation. 
As a result, the two gradients $\nabla_\tb{w} \tb{v}(S_t)$ and $\nabla_\tb{w} \tb{v}(S_{t+1}^\prime)$ tend to be similar and cancel each other, slowing down the learning. 
Theoretically, \citet{schoknecht2003td} prove TD converges faster than RG in a tabular setting. 

\item \citet{lagoudakis2003least} argue that TD usually provides a better solution than RG, even though the value function is not as well approximated. 
The TD solution ``preserves the shape of the value function to some extent rather than trying to fit the absolute values''. 
Thus ``the improved policy from the corresponding approximate value function is closer to the improved policy from the exact value function'' \citep{lagoudakis2003least,li2008worst,sun2015online}.
\end{itemize}

\textbf{Pros of RG:}
\begin{itemize}
\item RG is a true gradient algorithm and enjoys convergence guarantees in most settings under mild conditions. 
By contrast, the divergence of TD with off-policy learning or nonlinear function approximation is well documented \citep{tsitsiklis1997analysis}. Empirically, \citet{munos2003error} and \citet{li2008worst} show that RG is more stable than TD.
\item \citet{schoknecht2003td} observe that RG converges faster than TD in the four-room domain \citep{sutton1999between} with linear function approximation. \citet{scherrer2010should} shows empirically that the TD solution is usually slightly better than RG but in some cases fails dramatically. 
\end{itemize}

\textbf{Others:}
\begin{itemize}
\item \citet{li2008worst} proves that TD makes more accurate predictions (i.e., the predicted state value is close to the true state value), while RG yields smaller temporal differences (i.e., the value predictions for a state and its successor are more consistent). 
This is also explained in \citet{sutton2018reinforcement}.
\end{itemize}

To summarize, previous insights about RG and TD are mixed. There is little empirical study for RG in deep RL problems, much less RA. It is not clear whether and how we can take advantage of RA in model-free and model-based RL to solve deep RL problems.

\section{Residual Algorithms in Model-free RL}
\label{sec:mf}
In this section, we investigate how to combine RA and DDPG. 
In particular, we consider (almost) deterministic environments (e.g., DMControl) to avoid the double sampling issue.

In semi-gradient algorithms, value propagation goes backwards in time. 
The value of a state depends on the value of its successor through bootstrapping,
and a target network is used to stabilize this bootstrapping.
RA allows value propagation both forwards and backwards.
The value of a state depends on the value of both its successor and predecessor.
Therefore, we need to stabilize the bootstrapping in both directions.
To this end, we propose the \emph{bidirectional target network} technique. Employing this in DDPG yields Bi-Res-DDPG, which updates the critic parameters $\theta$ as:
\begin{align*}
\theta \leftarrow \theta &- \alpha_1 \big(r_{t+1} + \gamma \bar{Q}(s_{t+1}, \bar{\mu}(s_{t+1})) - Q(s_t, a_t) \big) \\
&\times \big(- \nabla_\theta Q(s_t, a_t)\big) \\
&- \alpha_1 \big( r_{t+1} + \gamma Q(s_{t+1}, \mu(s_{t+1})) - \bar{Q}(s_t, a_t) \big) \\
&\times \eta \gamma \nabla_\theta Q(s_{t+1}, \mu(s_{t+1})),
\end{align*}
where $\bar{Q}, \bar{\mu}$ are target networks and $\eta \in [0, 1]$ controls how the two gradients are mixed.  The actor update remains unchanged.

\begin{figure*}
\begin{center}
\includegraphics[width=0.8\textwidth]{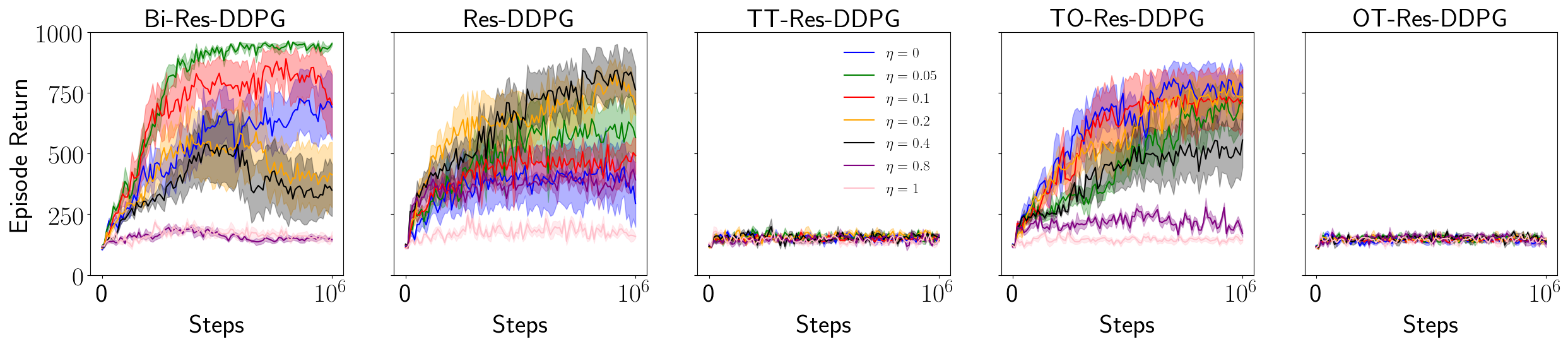}
\end{center}
\caption{\label{fig:ddpg-variants-mean} Performance of Bi-Res-DDPG variants on \texttt{walker-stand}, focusing on the role of target networks.}
\end{figure*}

\begin{figure}
\begin{center}
\includegraphics[angle=90,origin=c,width=0.4\textwidth]{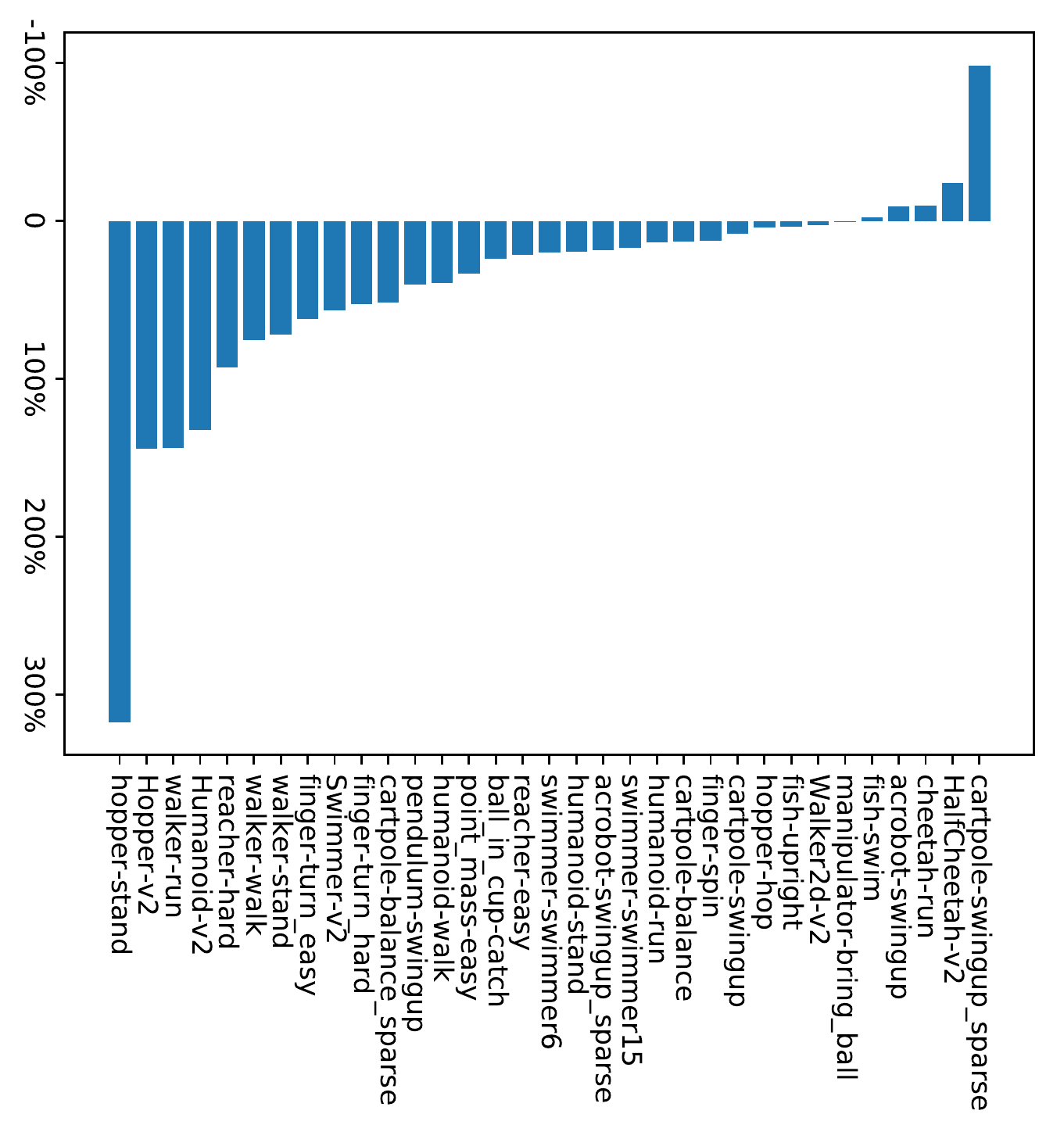}
\end{center}
\caption{\label{fig:ddpg-auc} AUC improvements of Bi-Res-DDPG over DDPG on 28 DMControl tasks and 5 Mujoco tasks, computed as $\frac{\text{AUC}_\text{Bi-Res-DDPG} - \text{AUC}_\text{DDPG}}{\text{AUC}_\text{DDPG}} $.}
\end{figure}

We compared Bi-Res-DDPG to DDPG in 28 DMControl tasks and 5 Mujoco tasks. 
Our DDPG implementation uses the same architecture and hyperparameters as \citet{lillicrap2015continuous}, 
which are inherited by Bi-Res-DDPG (and all other DDPG variants in this paper).
For Bi-Res-DDPG, we tune $\eta$ over $\{0, 0.05, 0.1, 0.2, 0.4, 0.8, 1\}$ on \texttt{walker-stand} and use $\eta=0.05$ across all tasks.
We perform 20 deterministic evaluation episodes every $10^4$ training steps and plot the averaged evaluation episode returns. 
All curves are averaged over 5 independent runs 
and are available in the appendix.
In the main text, 
we report the improvement of AUC (area under the curve) of the evaluation curves in Figure~\ref{fig:ddpg-auc}. 
AUC serves as a proxy for learning speed (e.g., see Example 8.2 in \citet{sutton2018reinforcement}).
Bi-Res-DDPG achieves a 20\% (41\%) AUC improvement over the original DDPG in terms of the median (mean). 
Our DDPG baseline reaches the same performance level as the DDPG baseline in \citet{fujimoto2018addressing} and \citet{buckman2018sample} in Mujoco tasks. 

To further investigate the relationship between the target network and RA, 
we study several variants of DDPG. 
We define a shorthand $g_t \doteq \eta\gamma \nabla_\theta Q(s_{t+1}, \mu(s_{t+1})) - \nabla_\theta Q(s_t, a_t)$ and the update rule for $\theta$ is $\theta \leftarrow \theta - \alpha_1 (r_{t+1} + \Delta) g_t$, 
where $\Delta$ is different for different variants.
We use ``T'' and ``O'' to denote the target network and the online network respectively. 
We have:
\begin{align}
\label{eq:res-ddpg}
\text{Res-DDPG:}& \Delta \doteq \gamma Q(s_{t+1}, \mu(s_{t+1})) - Q(s_t, a_t),\\
\text{TO-Res-DDPG:}& \Delta \doteq \gamma \bar{Q}(s_{t+1}, \bar{\mu}(s_{t+1})) - Q(s_t, a_t),\\
\text{OT-Res-DDPG:}& \Delta \doteq \gamma Q(s_{t+1}, \mu(s_{t+1})) - \bar{Q}(s_t, a_t),\\
\text{TT-Res-DDPG:}& \Delta \doteq \gamma \bar{Q}(s_{t+1}, \bar{\mu}(s_{t+1})) - \bar{Q}(s_t, a_t).
\end{align}
Res-DDPG is a direct combination of RA and DDPG without a target network.
TO-Res-DDPG simply adds a residual gradient term to the original DDPG.
OT-Res-DDPG stabilizes the bootstrapping for the forward value propagation.
TT-Res-DDPG stabilizes bootstrapping in both directions but destroys the connection between prediction and error.
By contrast, Bi-Res-DDPG stabilizes bootstrapping in both directions and maintains the connection between prediction and error.

\begin{figure}[t]
\begin{center}
\includegraphics[width=\linewidth]{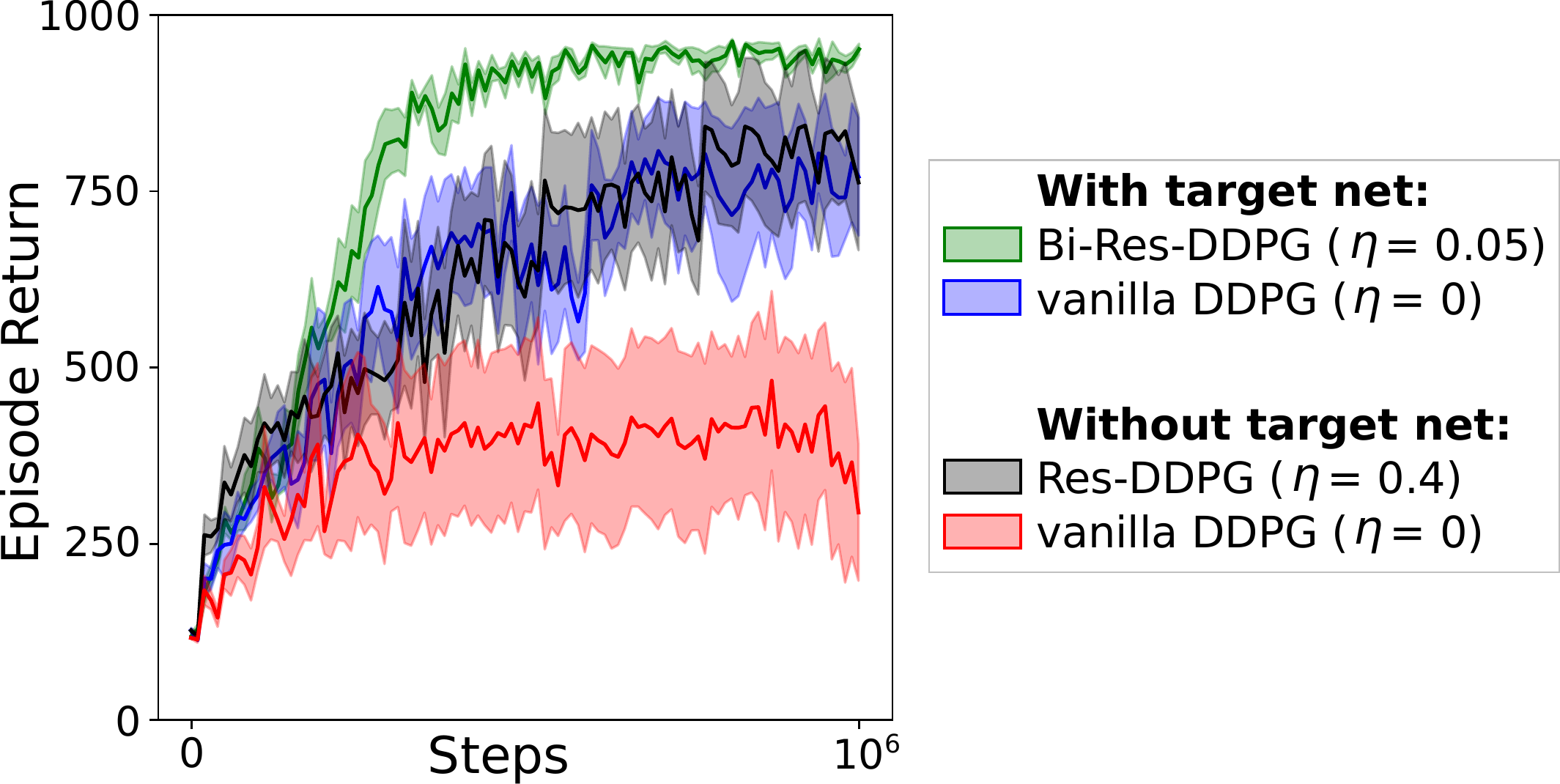}
\end{center}
\caption{\label{fig:ddpg-variants-summary} 
A selection of the best parameters $\eta$ from Figure~\ref{fig:ddpg-variants-mean}.
Note that residual updates stabilize performance as much as 
the introduction of target networks.
%A summary of Figure~\ref{fig:ddpg-variants-mean}, focusing on the role of residual updates.
}
\end{figure}

Figure~\ref{fig:ddpg-variants-mean} compares these variants on \texttt{walker-stand}.  The main points to note are:
(1) Both Bi-Res-DDPG($\eta=0$) and TO-Res-DDPG($\eta=0$) are the same as vanilla DDPG. 
The curves are similar, verifying the stability of our implementation. 
(2) Res-DDPG($\eta=0$) corresponds to vanilla DDPG without a target network, which performs poorly. 
This confirms that a target network is important for stabilizing training and mitigating divergence when a nonlinear function approximator is used \citep{mnih2015human,lillicrap2015continuous}.
(3) Increasing $\eta$ improves Res-DDPG's performance.
This complies with the argument from \citet{baird1995residual} that residual gradients help semi-gradients converge.
All variants fail with a large $\eta$ (e.g., 0.8 or 1).
This complies with the argument from \citet{baird1995residual} that pure residual gradients are slow.
(4) TO-Res-DDPG($\eta=0$) (i.e., vanilla DDPG) is similar to Res-DDPG($\eta=0.4$), indicating a naive combination of RA and DDPG without a target network is ineffective. 
(5) For TO-Res-DDPG, $\eta=0$ achieves the best performance, 
indicating adding a residual gradient term to DDPG directly is ineffective. 
To summarize, these variants confirm the necessity of the bidirectional target network.
To better understand the role of residual updates, 
we summarize the results of Figure~\ref{fig:ddpg-variants-mean} in Figure~\ref{fig:ddpg-variants-summary}.
Res-DDPG does not have a target network and outperforms DDPG without a target network. 
Res-DDPG also increases the stability.
Bi-Res-DDPG has target networks and also outperforms DDPG with a target network,
as well as increases the stability.
This comparison confirms the importance of residual updates.

We also evaluated a Bi-Res version of DQN in three ALE environments (BeamRider, Seaquest, Breakout). 
The performance was similar to the original DQN. 
One of the many differences between DMControl and ALE is that
rewards in ALE are much more sparse.
This might indicate that the forward value propagation in RA is less likely to yield a performance boost with sparse rewards.

We do not expect residual updates to improve the performance of all semi-gradient baselines. 
However, our results do show that the residual update together with the bidirectional target network is beneficial in many tasks.
Despite the popularity of semi-gradient methods, we do believe residual algorithms deserve more study.
The combination of residual updates and other semi-gradient algorithms, e.g., TD3 \citep{fujimoto2018addressing}, is a possibility for future work. 
We also do not address the double sampling issue in stochastic environments. 
This is indeed a restriction, 
but we would like to emphasize that
most available benchmarks with continuous
actions have deterministic transitions,
which indicates that this class of problems 
is of practical concern.

\section{Residual Algorithms in Model-based RL}

In model-based RL, the double sampling issue can be easily addressed by querying the learned model (either deterministic or stochastic).
Given the empirical success of deterministic models and their robustness in complex tasks \citep{kurutach2018model,feinberg2018model,buckman2018sample}, 
we consider deterministic models in this paper.
Dyna \citep{sutton1990integrated} is a commonly used model-based RL framework that trains a value function with imaginary transitions from a learned model. 
In this paper, we consider the combination of Dyna and DDPG.
For each planning step, we sample a transition $(s, a, r, \p{s})$ from a replay buffer and add some noise $\epsilon$ to the action $a$, 
yielding a new action $\hat{a}$.
We then query a learned model with $(s, \hat{a})$ and get $(\hat{r}, \p{\hat{s}})$.
This imaginary transition is then used to train the $Q$-function. 
The pseudocode of this Dyna-DDPG is provided in Algorithm~\ref{algo:dyna-ddpg}.
We aim to investigate different strategies for updating $Q$ during planning (i.e., the selection of $f$ in Algorithm~\ref{algo:dyna-ddpg}).

\begin{algorithm}[t]
\textbf{Input:} \;
$Q:$ a critic parameterized by $\theta$ \;
$\mu:$ an actor parameterized by $\nu$ \;
$P:$ planning steps \; 
$\epsilon:$ a noise process \;
$f:$ a critic update procedure \; \;
Initialize target networks $\theta^- \leftarrow \theta, \nu^- \leftarrow \nu$ \;
Initialize a replay buffer $\mathcal{B}$, a model $\mathcal{M}$ \;
Get an initial state $S_0$ and set $t \leftarrow 0$ \;
\While{true}{
  $A_t \leftarrow \mu(S_t)$ \;
  Execute $A_t$ and get $R_{t+1}, S_{t+1}$ \;
  Store $(S_t, A_t, R_{t+1}, S_{t+1})$ into $\mathcal{B}$ \; 
  Fit $\mathcal{M}$ with data in $\mathcal{B}$ \;
  Sample a \texttt{batch} of transitions from $\mathcal{B}$ \;
  \For{$(s, a, r, \p{s}) \, \textup{in} \, \textup{\texttt{batch}}$}{
    Update $\theta, \nu$ with $(s, a, r, \p{s})$, \eqref{eq:ddpg-critic}, \eqref{eq:ddpg-actor} \;
    \tcp{Planning}
    \For{$i \leftarrow 1, \dots, P$}{
      $\hat{a} \leftarrow a + \epsilon$ \;
      $\hat{r}, \p{\hat{s}} \leftarrow \mathcal{M}(s, \hat{a})$ \;
      Update $\theta$ with $(s, \hat{a}, \hat{r}, \p{\hat{s}})$ and $f$ \;
    }
  }
  $t \leftarrow t + 1$ \;
  Update $\theta^-, \nu^-$ according to $\theta, \nu$
}
\caption{\label{algo:dyna-ddpg} Dyna-DDPG}
\end{algorithm}

\begin{figure*}[h]
\begin{center}
\includegraphics[width=0.8\textwidth]{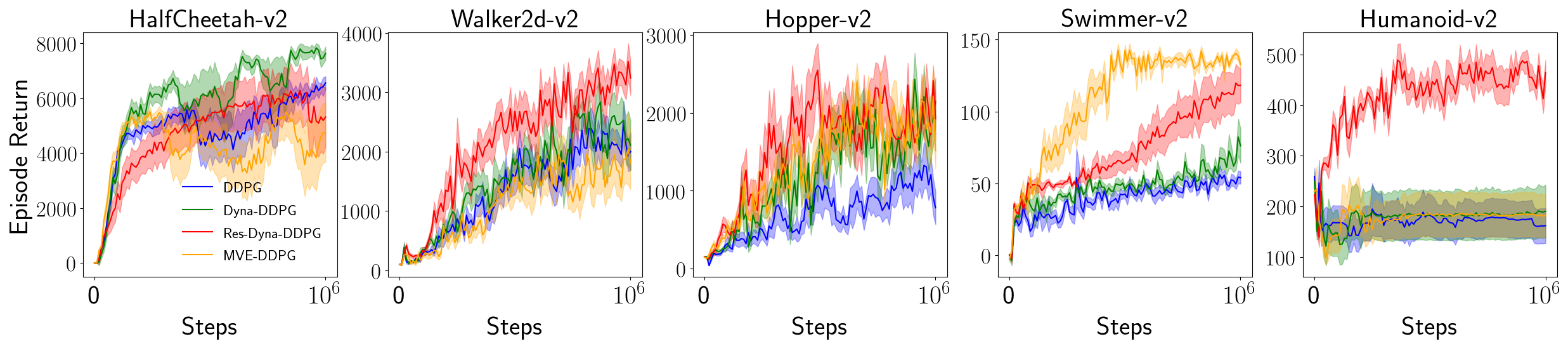}
\end{center}
\caption{\label{fig:ddpg-oracle-mean} Evaluation performance for different model-based DDPG with an oracle model.}
\end{figure*}

\begin{figure*}[h]
\begin{center}
\includegraphics[width=0.8\textwidth]{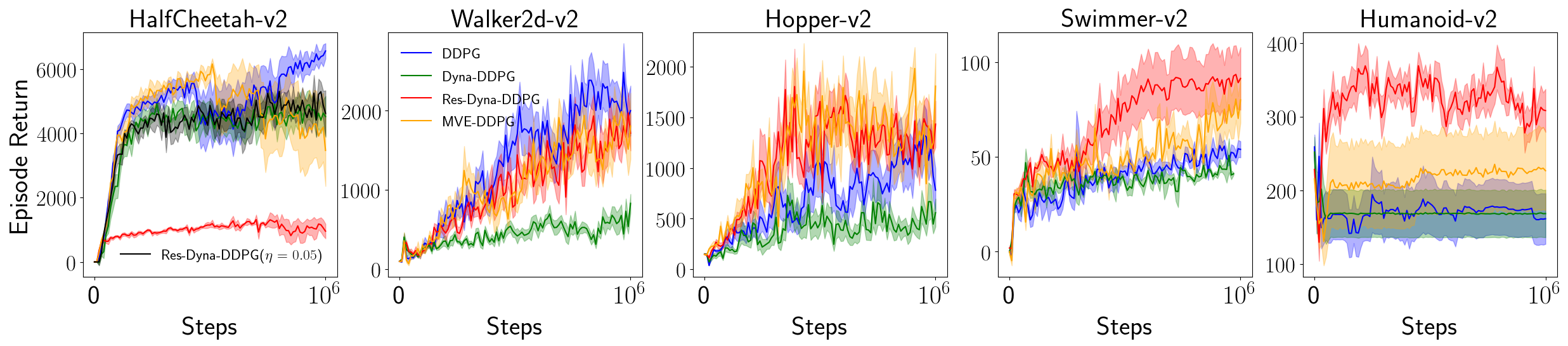}
\end{center}
\caption{\label{fig:ddpg-dyna-mean} Evaluation performance for different model-based DDPG with a learned model.}
\end{figure*}

One naive choice is to use the semi-gradient critic update \eqref{eq:ddpg-critic}. 
However, this suffers from the \emph{distribution mismatch} problem \citep{feinberg2018model}. 
When we apply \eqref{eq:ddpg-critic} in an imaginary transition $(s, \hat{a}, \hat{r}, \p{\hat{s}})$, we need the $Q$-value on $\p{\hat{s}}$ for bootstrapping. 
The $Q$-function is trained to make an accurate prediction on the state distribution of $s$, 
which is usually different from the state distribution of $\p{\hat{s}}$.  
This distribution mismatch results from both an imperfect model and the different sampling strategies for $a$ and $\hat{a}$.
It yields an inaccurate prediction for $Q(\p{\hat{s}}, \mu(\p{\hat{s}}))$, leading to poor performance \citep{feinberg2018model}. 
The TD($k$) trick \citep{feinberg2018model} is one attempt to address this issue. 
With a real transition $(s_{-1}, a_{-1}, r_0, s_0)$ sampled from a replay buffer, 
a model is unrolled for $k$ steps following $\bar{\mu}$, yielding a trajectory $(s_{-1}, a_{-1}, r_0, s_0, a_0,$ $r_1, s_1, \dots, r_k, s_k)$. 
TD($k$) then updates $\theta$ to minimize
\begin{align}
\label{eq:mve}
\textstyle
\nonumber
&\frac{1}{k+1} \sum_{t=-1}^{k-1} \Big( Q(s_t, a_t) \\
&- \big( \sum_{i=t+1}^k \gamma^{i-t-1}r_i + \gamma^{k-t}\bar{Q}(s_k, \bar{\mu}(s_k)) \big) \Big)^2.
\end{align}
With this update, $Q$ is trained on distributions of almost all the states ($s_{-1}, \dots, s_{k-1}$), which
\citet{feinberg2018model} show helps performance.
However, TD($k$) still does not train $Q$ on the last imaginary state $s_k$,
which is used for bootstrapping. 
On the one hand, the influence of the bootstrapping error from $s_k$ decreases as the trajectory gets longer thanks to discounting. 
On the other hand, even small state prediction errors typically compound as trajectories get longer, 
yielding a large prediction error of the state $s_k$ itself.
This contradiction is deeply embedded in TD($k$).
Consequently, TD($k$) must assume the model is accurate for $k$-step unrolling, which is
usually hard to satisfy in practice. 

In this paper, we seek to mitigate this distribution mismatch issue through RA.
For an imaginary transition $(s, \hat{a}, \hat{r}, \p{\hat{s}})$, RA naturally allows the $Q$-function to be trained on both $s$ and $\p{\hat{s}}$, 
without requiring further unrolling like TD($k$).
The use of RA in model-based planning is inspired by the theoretical results from \citet{li2008worst}, who
 proves that TD makes better predictions than RG.
On a real transition, this accelerates backward value propagation by providing better bootstrapping.
However, on an imaginary transition from a model, the value function is never trained on the imaginary successor state. 
It is questionable whether we should trust the value prediction on an imaginary state as much as a real state.
We, therefore, propose to use RA on imaginary transitions, which encourages the $Q$-function to be consistent with the model as showed by \citet{li2008worst}.

We now evaluate RA in model-based planning experimentally. We compare the performance of Dyna-DDPG($f=\text{Eq.}\eqref{eq:ddpg-critic}$) (referred to as Dyna-DDPG), Dyna-DDPG($f=\text{Eq.}\eqref{eq:res-ddpg}$) (referred to as Res-Dyna-DDPG), and DDPG+TD($k$) (referred to as MVE-DDPG following, \cite{feinberg2018model}).
We consider five Mujoco tasks used by \citet{buckman2018sample}, 
which is a superset of tasks used by \citet{feinberg2018model}.
In \citet{feinberg2018model}, the unrolling steps of MVE-DDPG are different for different tasks, which serve as  domain knowledge.
For a fair comparison, \citet{buckman2018sample} set $k=3$ for all tasks in their baseline MVE-DDPG. 
In our empirical study, we followed this convention.
We use a slightly different TD($k$) loss to improve stability of MVE-DDPG, which is explained in detail in the appendix.

To separate planning from model learning, we first consider planning with an oracle model.
In this section, we restrict our empirical study on Mujoco tasks as we do not have direct access to the oracle models in DMControl tasks.
We tune hyperparameters for Dyna-DDPG and Res-Dyna-DDPG on Walker and set $\eta = 0.2$ for all tasks. 
Other details are provided in the appendix. 
The results are reported in Figure~\ref{fig:ddpg-oracle-mean}.
Curves are averaged over 8 independent runs and shadowed regions indicate standard errors.
Both Dyna-DDPG and MVE-DDPG with an oracle model improve performance in 2 of 5 games, while Res-Dyna-DDPG improves performance in 4 out of 5 games.
These results suggest that RA is a more effective approach to exploit a model for planning. 
In HalfCheetah, both MVE-DDPG and Res-Dyna-DDPG fail to outperform Dyna-DDPG.
This could suggest that the distribution mismatch problem is not significant in this task. 
Furthermore, MVE-DDPG exhibits instability in HalfCheetah, which is also observed by \citet{buckman2018sample}.

We now consider planning with a learned model. We use the same model parameterization and model training protocol as \citet{feinberg2018model}. 
We set $\eta=0.2$ for all tasks.
The results are reported in Figure~\ref{fig:ddpg-dyna-mean}.
In Swimmer and Humanoid, Res-Dyna-DDPG significantly outperforms all other methods,
where Humanoid is usually considered to be the hardest task among all Mujoco tasks.
In Walker and Hopper, Res-Dyna-DDPG reaches similar performance as MVE-DDPG.
In HalfCheetah, Res-Dyna-DDPG ($\eta=0.2$) fails dramatically.
We further test other values for $\eta$ and find $\eta=0.05$ produces reasonable performance, 
as shown by the extra black curve.
This indicates that $\eta$ can serve as domain knowledge, reflecting our confidence in a learned model. 
A possibility for future work is to use model uncertainty estimation from a model ensemble to determine $\eta$ automatically, 
similar to what \citet{buckman2018sample} propose for the unrolling steps in TD($k$),
which significantly improves performance over MVE-DDPG.

In this section, we consider the vanilla residual update \eqref{eq:res-ddpg} without the bidirectional target network. 
Our preliminary experiments show that introducing the bidirectional target network during planning does not further boost performance.
The main purpose of a target network is to stabilize bootstrapping (value propagation).
Due to the distribution mismatch problem on imaginary transitions, however,
it may be more important for the value function to be consistent with the model than simply propagating the value in either direction.
This may reduce the importance of the bidirectional target network.

\begin{figure*}[h]
\begin{center}
\includegraphics[width=0.8\textwidth]{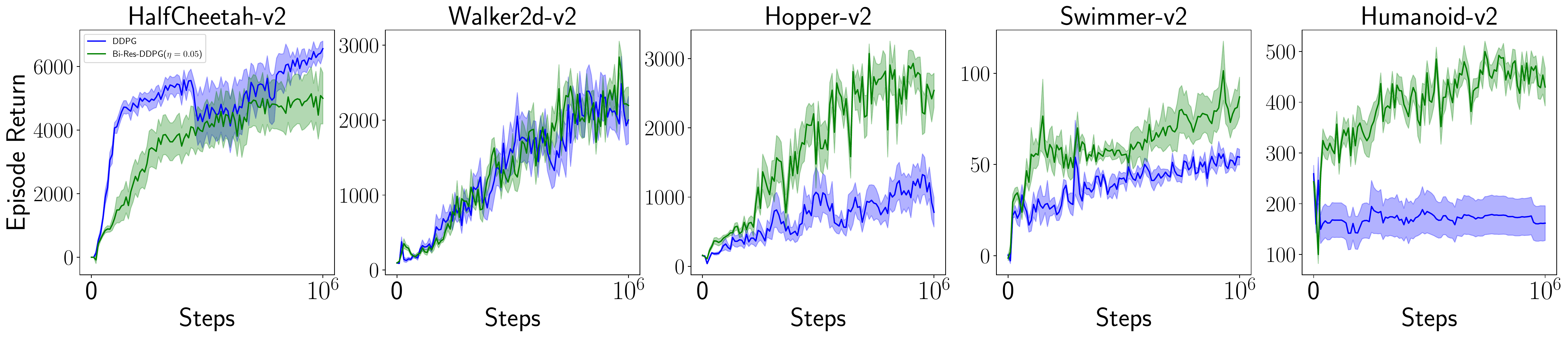}
\end{center}
\caption{\label{fig:ddpg-mf-mujoco-mean} Evaluation curves of DDPG and Bi-Res-DDPG($\eta=0.05$) on 5 Mujoco tasks. Curves are averaged over 5 independent runs and shaded regions indicate standard errors.}
\end{figure*}

\section{Related Work}

There are other studies on Bellman residual methods. \citet{geist2017bellman} show that for policy-based methods, maximizing the average reward is better than minimizing the Bellman residual. 
\citet{schoknecht2003convergent} show RG converges with a problem-dependent constant learning rate when combined with certain function approximators.
\citet{dabney2014natural} extend RG with natural gradients.
However, this paper appears to be the first to contrast residual gradients and semi-gradients in deep RL problems and demonstrate the efficacy of RA with new algorithms.
\citet{dai2017sbeed} attack the double sampling issue via dual embedding, 
which translates the minimization of MSBE into a minimax problem.
For this translation to hold, the maximization step has to be conducted over a function class which is rich enough to contain the true maximizer.
This condition, however, does not necessarily hold when linear function approximation is considered.
It can be easily verified that with linear function approximation, dual embedding indeed translates MSBE into MSPBE. 
Besides MSPBE, other losses have also been proposed to avoid the double sampling issue in minimizing MSBE,
for example, \citet{feng2019kernel} propose a kernel loss based on the Bellman equation, \citet{antos2008learning} add a penalty term to MSBE.

Dyna-style planning in RL has been widely used. 
\citet{gu2016continuous} learn a local linear model for planning. 
\citet{kurutach2018model} learn a model ensemble to avoid overfitting to an imperfect model, which is also achieved by meta-learning \citep{clavera2018model}. 
\citet{kalweit2017uncertainty} use a value function ensemble to decide when to use a model.
Besides Dyna-style planning, learned models are also used for a lookahead tree-search to improve value estimation at decision time \citep{silver2017predictron,oh2017value,talvitie2017self}. 
This tree-search is also used as an effective inductive bias in value function parameterization \citep{farquhar2018treeqn,srinivas2018universal,zhang2018ace}.
Trajectories from a learned model are also used as extra inputs for value functions \citep{weber2017imagination}, which reduces the negative influence of the model prediction error.
In this paper, we focus on the simplest Dyna-style planning and leave the combination of RA and more advanced planning techniques for future work.

Besides RL, learned models are also used in other control methods, e.g., model predictive control (MPC, \cite{garcia1989model}). 
\citet{nagabandi2018neural} learn deterministic models via neural networks for MPC.
\citet{chua2018deep} conduct a thorough comparison between deterministic models and stochastic models and use particle filters when unrolling a model.
Besides modeling the observation transition, 
\citet{ha2018world} and \citet{hafner2018learning} propose to model the abstract state transition and use MPC on the abstract state space. 
In this paper, we focus on the simplest deterministic model and leave the combination of RA and more advanced models for future work.

\section{Conclusions}
In this paper, we give a thorough review of existing comparisons between RG and TD. 
We propose the bidirectional target network technique to stabilize bootstrapping in both directions in RA, yielding a significant performance boost. 
We also demonstrate that RA is a more effective approach to the distribution mismatch problem in model-based planning than the existing TD($k$) method.
Our empirical study showed the efficacy of RA in deep RL problems, 
which has long been underestimated by the community.
A possibility for future work is to study RA in model-free RL with stochastic environments, 
where the double sampling issue cannot be trivially resolved.

\begin{acks}
SZ is generously funded by the Engineering and Physical Sciences Research Council (EPSRC). This project has received funding from the European Research Council under the European Union's Horizon 2020 research and innovation programme (grant agreement number 637713). The experiments were made possible by a generous equipment grant from NVIDIA.
\end{acks}

\appendix

\section{Experiment Details}
All our implementations and the corresponding Docker environment are made publicly available.\footnote{\url{https://github.com/ShangtongZhang/DeepRL}} 
Open AI Gym and DMControl are available at~\url{https://gym.openai.com/} and~\url{https://github.com/deepmind/dm_control}.

Our DDPG implementation uses the same parameterization and hyperparameters as \citet{lillicrap2015continuous},
which are inherited by all the variants of DDPG in this paper without further turning.
We do not use batch normalization. 

For the model-based experiments,
we tune extra hyperparameters in Walker with an oracle model for both Dyna-DDPG and Res-Dyna-DDPG.
The planning steps $P$ is tuned over $\{1, 2, 4\}$. 
The noise process $\epsilon$ is Gaussian noise $\mathcal{N}(0, \sigma^2)$, with $\sigma$ tuned over $\{0.05, 0.1, 0.2\}$. 
The mix coefficient $\eta$ in RA is tuned over \\$\{0, 0.05, 0.1, 0.2, 0.4, 0.8, 1\}$.
In all our experiments (with both an oracle model and a learned model), we set $P=1, \sigma=0.1, \eta=0.2$. 

For MVE-DDPG, we find the original TD($k$) loss~\eqref{eq:mve} yields significant instability. 
To improve stability, we made two modifications.
First, for a trajectory $(s_{-1}, a_{-1}, r_0, s_0, a_0, r_1, s_1, \dots, r_k, s_k)$, instead of minimizing the loss~\eqref{eq:mve}, we minimize
\begin{align*}
&\Big( Q(s_{-1}, a_{-1}) - \big(r_0 + \gamma \bar{Q}(s_0, a_0) \big)  \Big)^2 \\
&+ \frac{1}{k} \sum_{t=0}^{k-1} \Big( Q(s_t, a_t) - \big( \sum_{i=t+1}^k \gamma^{i-t-1}r_i + \gamma^{k-t}\bar{Q}(s_k, \bar{\mu}(s_k)) \big) \Big)^2.
\end{align*}
This new loss is different from \eqref{eq:mve} mainly in that it uses the real transition $(s_{-1}, a_{-1}, r_0, s_0)$ more. 
We find this significantly improves stability.
Second, we replace the mean squared loss with a Huber loss \citep{huber1964robust},
which has been reported to improve stability \citep{baselines}.
Our MVE-DDPG implementation significantly outperforms the MVE-DDPG baselines in \citet{buckman2018sample} in Hopper and Walker 
while maintains a similar performance in remaining tasks.
The MVE-DDPG in \citet{feinberg2018model} has task-dependent learning rates.
By contrast, we do not tune hyperparameters task by task for any compared algorithm.

We conducted our experiments on an Nvidia DGX-1 with PyTorch.

\section{Other Experimental Results}
\begin{figure*}
\begin{center}
\includegraphics[width=0.7\textwidth]{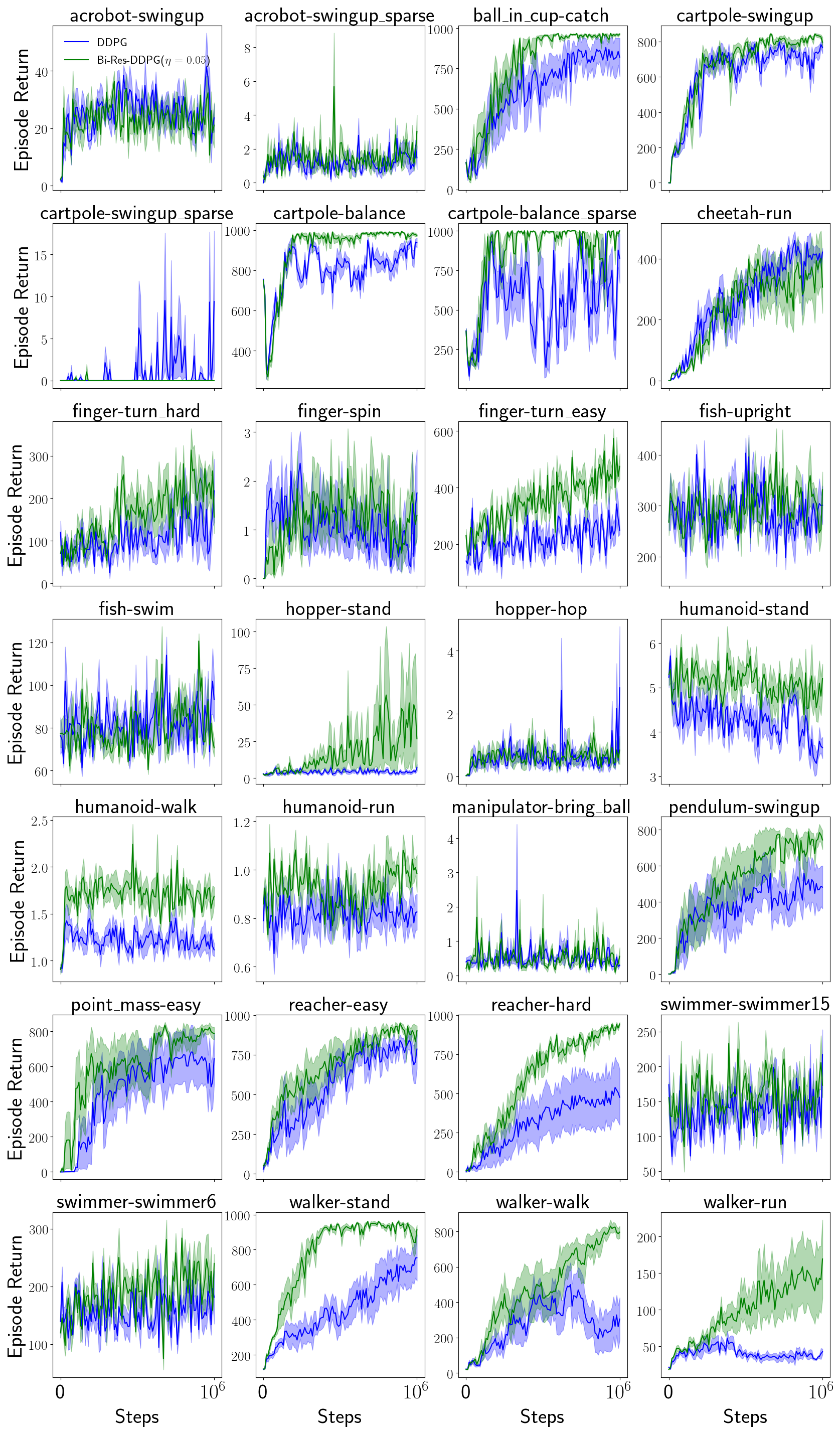}
\end{center}
\caption{\label{fig:ddpg-mf-mean} Evaluation curves of DDPG and Bi-Res-DDPG($\eta=0.05$) on 28 DMControl tasks. Curves are averaged over 5 independent runs and shaded regions indicate standard errors.}
\end{figure*}

The evaluation curves of DDPG and Bi-Res-DDPG($\eta=0.05$) on 5 Mujoco tasks and 28 DMControl tasks are reported in Figure~\ref{fig:ddpg-mf-mujoco-mean} and Figure~\ref{fig:ddpg-mf-mean} respectively.

%%%%%%%%%%%%%%%%%%%%%%%%%%%%%%%%%%%%%%%%%%%%%%%%%%%%%%%%%%%%%%%%%%%%%%%%%%%%%%%%%%%%%%%%%%%%%%%%%%%%%%%%%
%% bibliography: see CFP for number of permitted pages

\bibliographystyle{ACM-Reference-Format}  % do not change this line!
\bibliography{ref}  % put name of your .bib file here

\end{document}